\documentclass[]{nature}
\pdfoutput=1
\usepackage{graphicx}
\DeclareGraphicsExtensions{.pdf,.png,.jpg}
\usepackage{amsmath} 
\usepackage{color} 
\makeatletter

\definecolor{lightgray}{rgb}{0.8,0.8,0.8}
\definecolor{darkblue}{rgb}{0,0,0.5}

\begin{document}

\newcommand{\transpose}[1]{#1^T}

\newcommand{\cameraImage}{\mathbf{I}}
\newcommand{\measuredImage}{\mathbf{M}}
\newcommand{\backgroundImage}{\mathbf{B}}
\newcommand{\ambientImage}{\mathbf{A}}
\newcommand{\objectImage}{\mathbf{O}}
\newcommand{\noiseImage}{\mathbf{N}}
\newcommand{\simulatedImage}[1]{\mathbf{S}\left(#1\right)}
\newcommand{\parameter}{p}

\newcommand{\norm}[1]{\left\Vert #1\right\Vert _{2}}
\newcommand{\normalize}[1]{#1}
\newcommand{\dotProduct}[2]{\left<#1,~#2\right>}
\newcommand{\normalizedDotProduct}[2]{#1\circ#2}

\newcommand{\genericVectorA}{v}
\newcommand{\genericVectorB}{w}

\newcommand{\todo}[2][]{{\textcolor{red}{\bf TODO: #2}}} 
\newcommand{\final}[1]{#1}

\title{Optically lightweight tracking of objects around a corner
\author{Jonathan Klein$^{2,1}$, Christoph Peters$^{1}$, Jaime Mart\'in$^{1}$, Martin Laurenzis$^{2}$ \& Matthias B. Hullin$^{1,}\footnote{
Corresponding author. \\Matthias B. Hullin\\University of Bonn\\Institute for Computer Science II\\Friedrich-Ebert-Allee 144\\53113 Bonn, Germany\\e-mail: hullin@cs.uni-bonn.de
}$}
}

\maketitle
\final{
\begin{affiliations}
 \item University of Bonn, Germany
 \item French-German Research Institute Saint-Louis, France
\end{affiliations}

}


\begin{abstract}
The observation of objects located in inaccessible regions is a recurring challenge in a wide variety of important applications.
Recent work has shown that indirect diffuse light reflections can be used to reconstruct objects and two-dimensional (2D) patterns around a corner. However, these prior methods always require some specialized setup involving either ultrafast detectors or narrowband light sources. Here we show that occluded objects can be tracked in real time using a standard 2D camera and a laser pointer. Unlike previous methods based on the backprojection approach, we formulate the problem in an analysis-by-synthesis sense. By repeatedly simulating light transport through the scene, we determine the set of object parameters that most closely fits the measured intensity distribution. We experimentally demonstrate that this approach is capable of following the translation of unknown objects, and translation and orientation of a known object, in real time. 
\end{abstract}


The widespread availability of digital image sensors, along with advanced computational methods, has spawned new imaging techniques that enable seemingly impossible tasks. A particularly fascinating result is the use of ultrafast time-of-flight measurements\cite{Abramson:78,Velten:2011} to image objects outside the direct line of sight\cite{Velten:2012:Recovering,Heide:2014,Laurenzis:2014,Gariepy:2016}. Being able to use arbitrary walls as though they were mirrors can provide a critical advantage in many sensing scenarios with limited visibility, like endoscopic imaging, automotive safety, industrial inspection and search-and-rescue operations.

\begin{figure}
\centering
\includegraphics[width=\linewidth,clip=0mm 0 0 0]{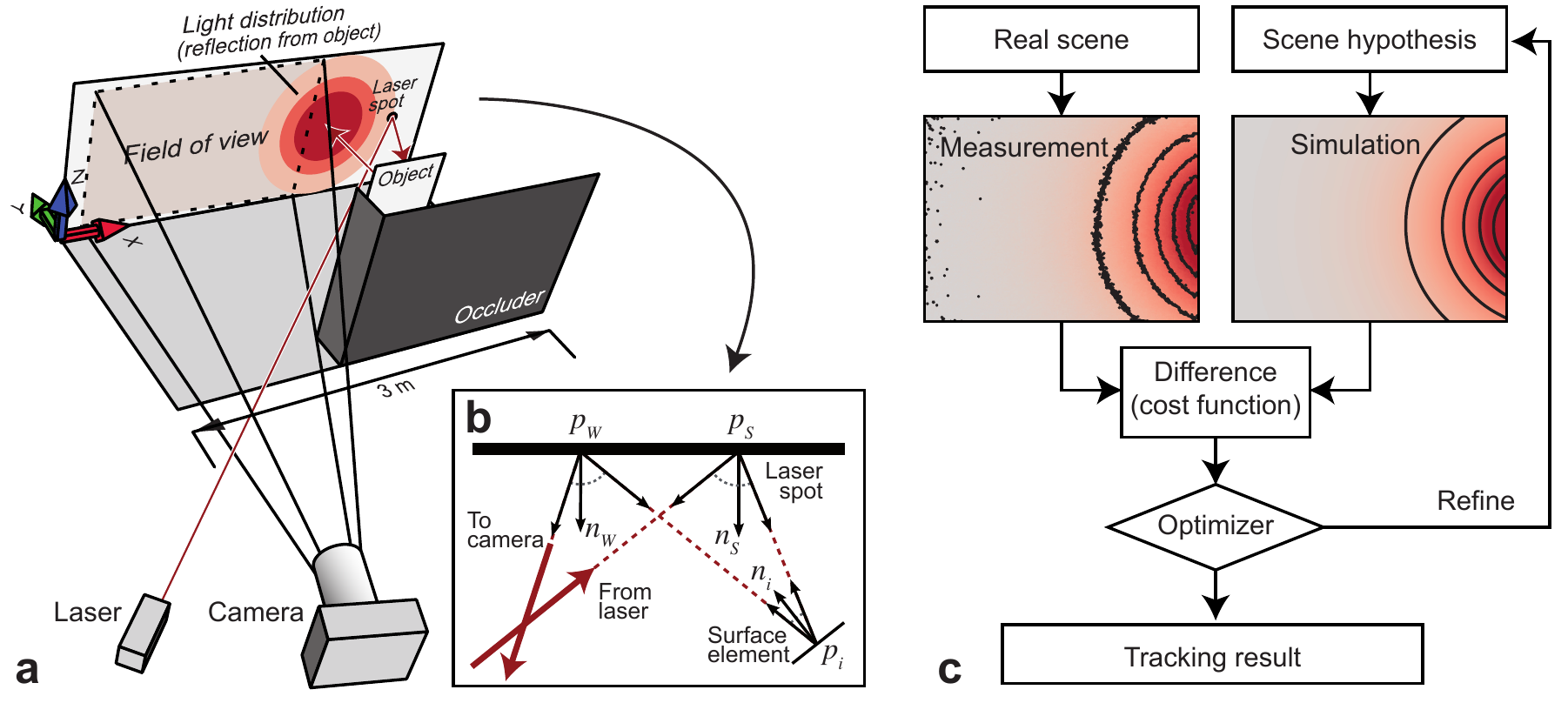}
\caption{
\label{fig:Overview}
Tracking objects around a corner. {\bf\sffamily a}, Our experimental setup follows the most common arrangement reported in prior work, except that it does not use time-of-flight technology. A camera observes a portion of a white wall. To the right of the camera's field of view, a collimated laser illuminates a spot that reflects light toward the unknown object. The light distribution observed by the camera is the result of three diffuse light bounces (wall--object--wall) plus ambient contributions.
{\bf\sffamily b}, Geometry of three-bounce reflection for a single surface element.
{\bf\sffamily c}, Flow diagram of our tracking algorithm. Given shape, position and orientation of an object (the ``scene hypothesis''), we simulate light transport to predict the distribution that this object would produce on the wall. By comparing this distribution to the one actually observed by the camera, and refining the parameters to minimize the difference, the object's motion is estimated.
}
\end{figure}

Out of the proposed techniques for imaging occluded objects, some require the object to be directly visible to a structured\cite{Sen:2005:DP} or narrow-band\cite{Steinvall:2011,katz2012looking,Katz:2014} light source.
Others resort to alternative regions in the electromagnetic spectrum where the occluder is transparent\cite{Sume:2011,Adib:2013,Adib:2015}. 
We adopt the more challenging assumption that the object is in the direct line of sight of neither light source nor camera (Fig.~\ref{fig:Overview}), and that it can only be illuminated or observed indirectly via a diffuse wall\cite{Velten:2012:Recovering,Heide:2014,Laurenzis:2014,Buttafava:2015,Gariepy:2016}. All the observed light has undergone at least three diffuse reflections (wall, object, wall), and reconstructing the unknown object is an ill-posed inverse problem
. Most solution approaches reported so far use a backprojection scheme as in computed tomography\cite{Pan:2009}, where each intensity measurement taken by the imager votes for a manifold of possible scattering locations. This \emph{explicit} reconstruction scheme is computationally efficient, in principle real-time capable\cite{Gariepy:2016}, and can be extended with problem-specific filters \cite{Velten:2012:Recovering,Kadambi:2016:OIT:2882845.2836164}. However, it assumes the availability of ultrafast time-resolved optical impulse responses, whose capture still constitutes a significant technical challenge. Techniques proposed in literature include direct temporal sampling based on holography\cite{Abramson:78,Abramson:83:LIF,Quercioli:85}, streak imagers\cite{Velten:2011}, gated image intensifiers\cite{Laurenzis:2014}, serial time-encoded amplified microscopy\cite{goda2009}, single-photon avalanche diodes\cite{gariepy2015single}, and indirect computational approaches using multi-frequency lock-in measurements\cite{Heide:2013:LBT,kadambi2013coded,peters-2015-fast-transient}.
In contrast, \emph{implicit} methods state the reconstruction task in terms of a problem-specific cost function that measures the agreement of a scene hypothesis with the observed data and additional model priors. The solution to the problem is defined as the function argument that minimizes the cost. In the only such method reported so far\cite{Heide:2014}, the authors regularize a least-squares data term with a computationally expensive sparsity prior, which enables the reconstruction of unknown objects around a corner without the need for ultrafast light sources and detectors.

Here  we introduce an implicit technique for detecting and tracking objects outside the line of sight in real time. Imaged using routinely available hardware (2D camera, laser pointer), the low-frequent intensity distribution reflecting off the object to the wall serves as our main source of information. To this end, we combine a simulator for three-bounce indirect light transport with a reduced formulation of the reconstruction task. Rather than aiming to reconstruct the geometry of an unknown object, we assume that the target object is rigid, and that its shape and material are either known and/or irrelevant. Translation and rotation, the only remaining degrees of freedom, can now be found by minimizing a least-squares energy functional, forcing the scene hypothesis into agreement with the captured intensity image. 

Our main contributions are threefold. We propose to use light transport simulation to tackle an indirect vision task in an analysis-by-synthesis sense. Using synthetic measurements, we quantify the effect of object movement on the observed intensity distribution, and predict under which conditions the effect is significant enough to be detected. Finally, we demonstrate and evaluate a hardware implementation of a tracking system. Our insights are not limited to intensity imaging, and we believe that they will bring non-line-of-sight sensing closer to practical applications.


\section*{Results}

\paragraph{Light transport simulation (synthesis). } At the center of this work is an efficient renderer for three-bounce light transport. Being able to simulate indirect illumination at an extremely fast rate is crucial to the overall system performance, since each object tracking step requires multiple simulation runs. Like all prior work, we assume that the wall is planar and known, and so is the position of the laser spot. The object is represented as a collection of Lambertian surface elements (\emph{surfels}), each characterized by its position, normal direction and area. As the object is moved or rotated, all its surfels undergo the same rigid transformation. We represent this transformation by the \emph{scene parameter} $\parameter$, which is a three-dimensional vector for pure translation, or a six-dimensional vector for translation and rotation. The irradiance received by a given camera pixel is computed by summing the light that reflects off the surfels. The individual contributions, in turn, are obtained independently of each other as detailed in the Methods section, by calculating the radiative transfer from the laser spot via a surfel to the location on the wall observed by a pixel.
Note that by following this procedure, like all prior work, we neglect self-occlusion, occlusion of ambient light, and interreflections. 
To efficiently obtain a full-frame image, represented by the vector of pixel values $\simulatedImage{\parameter}$, we parallelized the simulation to compute each pixel in a separate thread on the graphics card. The rendering time is approximately linear in the number of pixels and the number of surfels. On an NVIDIA GeForce GTX 780 graphics card, the response from a moderately complex object (500 surfels) at a resolution of 160$\times$128 pixels is rendered in 3.57 milliseconds. 

\begin{figure}
\centering
\includegraphics[width=\linewidth]{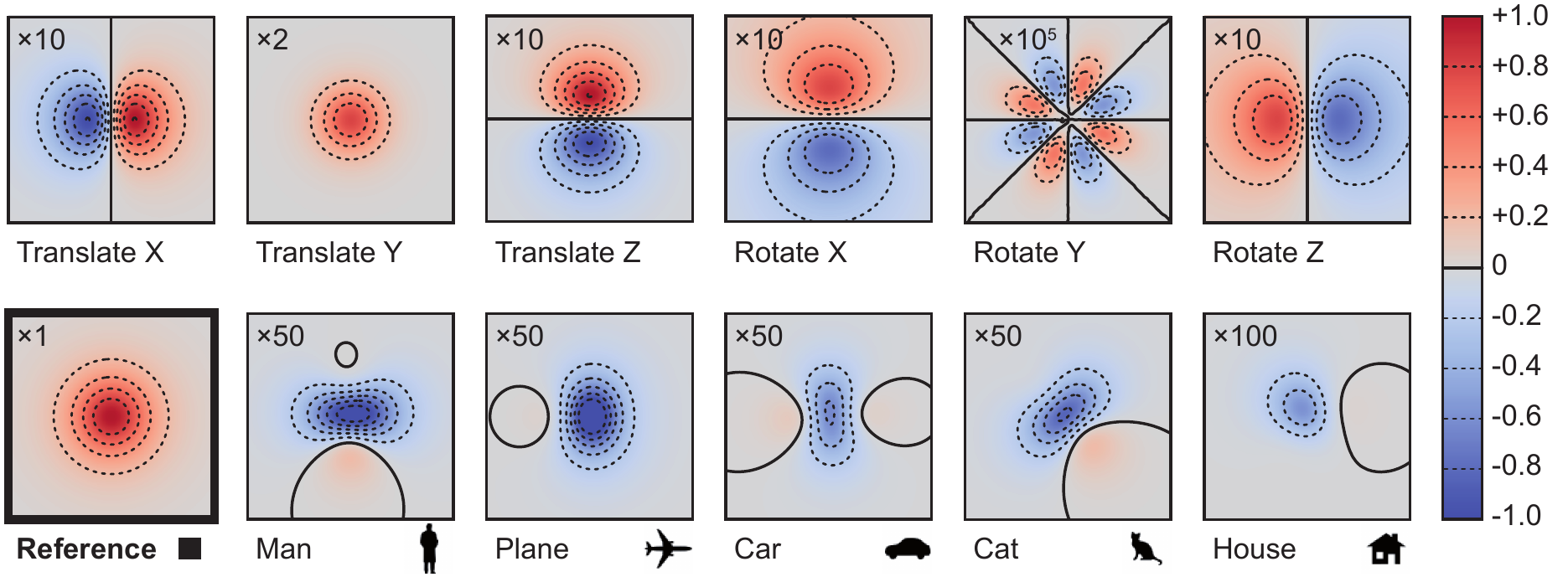}
\caption{
\label{fig:DifferenceImages}
Intensity difference images. To investigate the effect of changes in object position and orientation on the intensity distribution observed on the wall, we performed a simplified synthetic experiment with an orthographic view of a 2\,m$\times$2\,m wall, and laser spot and object  centered with respect to the wall. The reference distribution (bottom left) was produced by a 10\,cm$\times$10\,cm square-shaped object, located at 50\,cm from the wall. Six difference images (top row), obtained by translating ($\pm$2.5\,cm) and rotating ($\pm$7.5$^\circ$) the object about the X, Y and Z axes, illustrate the distribution and magnitude of the respective change in the signal. The  images shown in the bottom row visualize the difference caused by a change in shape. For display, each difference image has been amplified by the indicated factor (2 to 100,000) that also reflects the relative significance of the effect: Translations and rotations (except around the Y axis) caused the signal to change by roughly 1\% per centimeter or per angular degree. A change in the object shape led to a peak difference around 1--2\%, and rotation around the Y axis had a much smaller effect.
}
\end{figure}

To estimate the magnitude of changes in the intensity distribution that are caused by motion or a change in shape, we performed a numerical experiment using this simulation. In this experiment, we used a fronto-parallel view on a 2\,m$\times$2\,m wall, with a small planar object (a 10\,cm$\times$10\,cm white square) located at 50\,cm from the wall. Object and laser spot were centered on the wall, but not rendered into the image.
Fig.~\ref{fig:DifferenceImages} shows the simulated response thus obtained. By varying position and location of the object, we obtained difference images that can be interpreted as partial derivatives with respect to the components of the scene parameter $p$. Since the overall light throughput drops with the fourth power of the object-wall distance%
, translation in Y direction caused the strongest change. Translation in all directions and rotation about the X and Z axes affected the signal more strongly than the other variations. With differences amounting to several percent of the overall intensity, these changes were significant enough to be detected using a standard digital camera with 8- to 12-bit A/D converter. 

\paragraph{Experimental setup. } Our experiment draws inspiration from prior work\cite{Velten:2012:Recovering,Heide:2014,Buttafava:2015,Gariepy:2016,Kadambi:2016:OIT:2882845.2836164}; the setup is sketched in Fig.~\ref{fig:Overview}(a). Here, due to practical constraints, some of the idealizing assumptions made during the synthetic experiment had to be relaxed. In particular, only an off-peak portion of the intensity pattern could be observed, and we had to shield the camera from the laser spot to avoid lens flare. The actual reflectance distribution of the wall and object surfaces was not perfectly Lambertian, and additional light emitters and reflectors, not accounted for by the simulation, were present in the scene. 
To obtain a measured image $\measuredImage$ containing only light from the laser, we took the difference of images captured with and without laser illumination. Additionally, we subtracted a calibration measurement $\widehat\backgroundImage$ containing light reflected by the background. A specification of the devices used, and a more detailed introduction of the data pre-processing steps, can be found in the Methods section.

\paragraph{Tracking algorithm (analysis). } With the light transport simulation at hand, and given a measurement of light scattered from the object to the wall, we formulate the tracking task as a non-linear minimization problem. Suppose $\measuredImage$ and $\simulatedImage{\parameter}$ are vectors encoding the pixel values of the \emph{measured} object term and the one predicted by the \emph{simulation} under the transformation parameter or scene hypothesis $\parameter$, respectively. We search for the parameter $\parameter$ that brings $\measuredImage$ and $\simulatedImage{\parameter}$ into the best possible agreement by minimizing the cost function
\begin{equation}
f(\parameter)=\norm{\measuredImage-\gamma\left(\measuredImage,\simulatedImage{\parameter}\right)\cdot\simulatedImage{\parameter}}^2\mbox{,}\qquad\textrm{ where }\qquad\gamma(a,b)=\frac{\transpose{a}\cdot b}{\norm{b}^2}\mbox{ .}
\label{eq:objective}
\end{equation} 
The factor $\gamma(a,b)$ projects $b$ to $a$, minimizing the distance $\norm{a-\gamma(a,b)\cdot b}^2$. By including this factor into our objective, we decouple the recovery of the scene parameter $\parameter$ from any unknown 
global scaling between measurement and simulation, caused by parameters such as surface albedos, camera sensitivity and laser power. 
To solve this non-linear, non-convex, heavily over-determined problem, we use the Levenberg-Marquardt algorithm\cite{marquardt1963algorithm} as implemented in the Ceres library\cite{ceres-solver}. Derivatives are computed by numerical differentiation. When tracking six degrees of freedom (translation and rotation), evaluating the value and gradient of $f$ requires a total of seven simulation runs, or on the order of 25 milliseconds of compute time on our system.

\paragraph{Tracking result. }

\begin{figure}
\centering
\includegraphics[width=150mm]{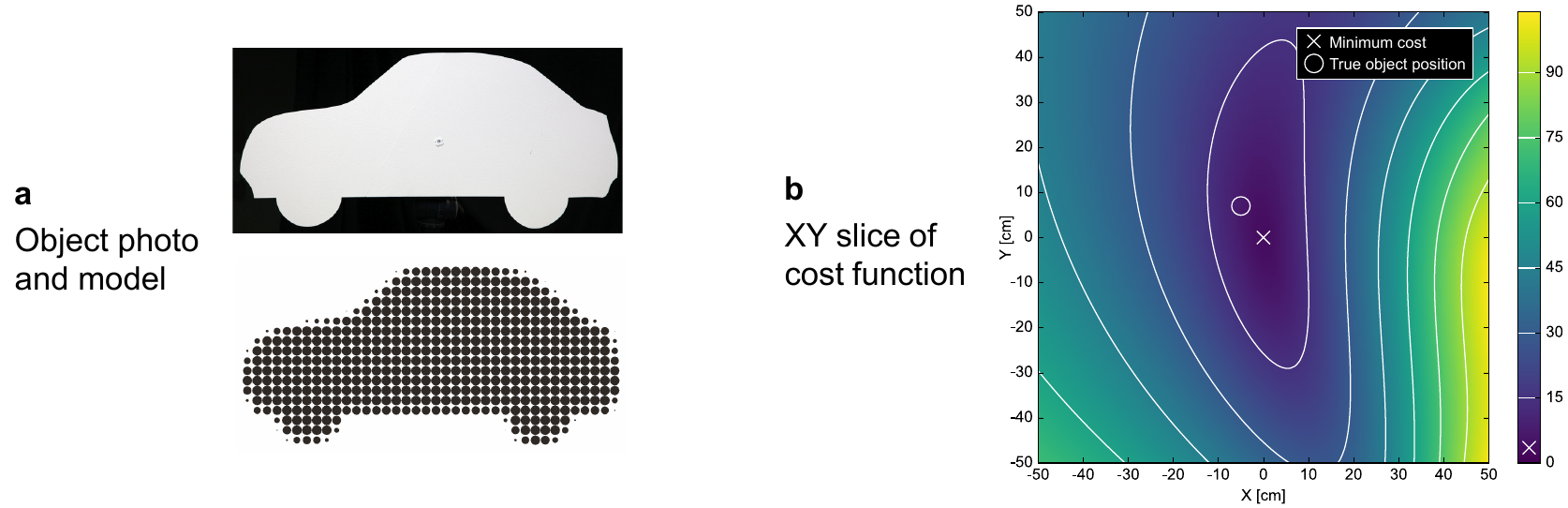}
\caption{
\label{fig:ObjectShape}
Object model and cost function used for tracking. {\bf\sffamily a,} photo of an object cut from white plywood, and its representation as surface elements (surfels). Note that although we use a flat object for demonstration, our method is also capable of handling three-dimensional objects.
{\bf\sffamily b,} XY slice of the cost function for positional tracking, centered around the global minimum. With a perfect image formation model and in the absence of noise, the minimum (marked by cross) and the measured position of the object (marked by circle) should coincide at a function value of exactly $f(p)= 0$. Under real conditions, the reconstructed position deviated from the true one by a few centimeters, and the minimum was a small positive value. 
}
\end{figure}

\begin{figure}
\centering
\includegraphics[width=\linewidth]{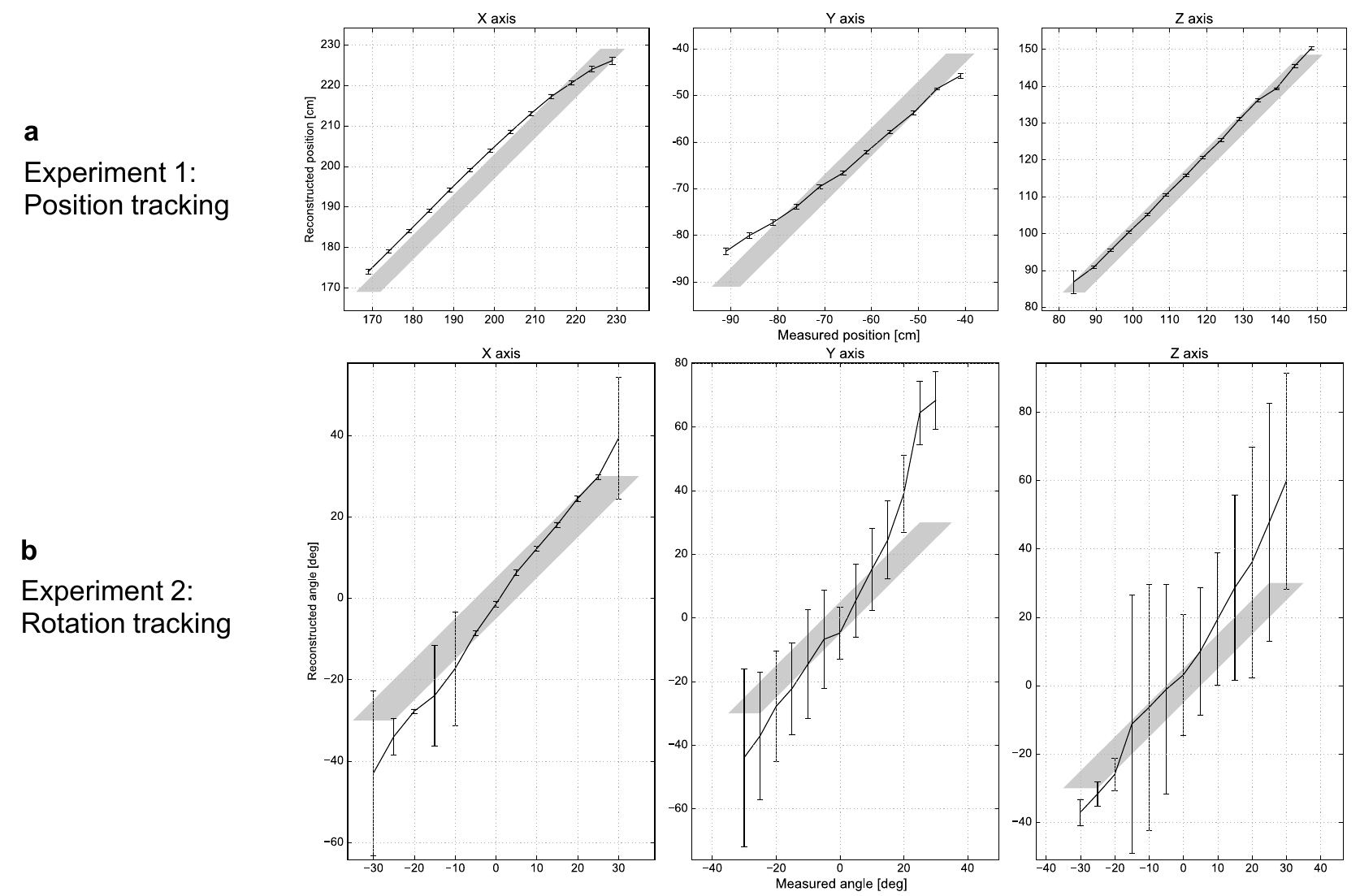}
\caption{
\label{fig:KnownObjectTracking}
Tracking a known object. {\bf\sffamily a,} Result of three tracking sessions where the object was translated along the X, Y and Z axes (Experiment 1). We recorded 100 input images at each position and reconstructed the object position for each input image independently. Plots and error bars visualize the mean and standard deviation of the recovered positions. The area shaded in gray is the confidence range for the true position which was determined using a tape measure.
{\bf\sffamily b,} Result of three tracking sessions where the object was rotated around the X, Y and Z axes (Experiment 2). From 100 input images, we jointly reconstructed translation and rotation. Shown are mean and standard deviation of the recovered rotation angle. The higher uncertainty reflects the fact that rotation in general has a smaller effect on the signal, and the ambiguity between translational and rotational motion (also see Fig.~2.).
}
\end{figure}

\begin{figure}
\centering
\includegraphics[width=\linewidth]{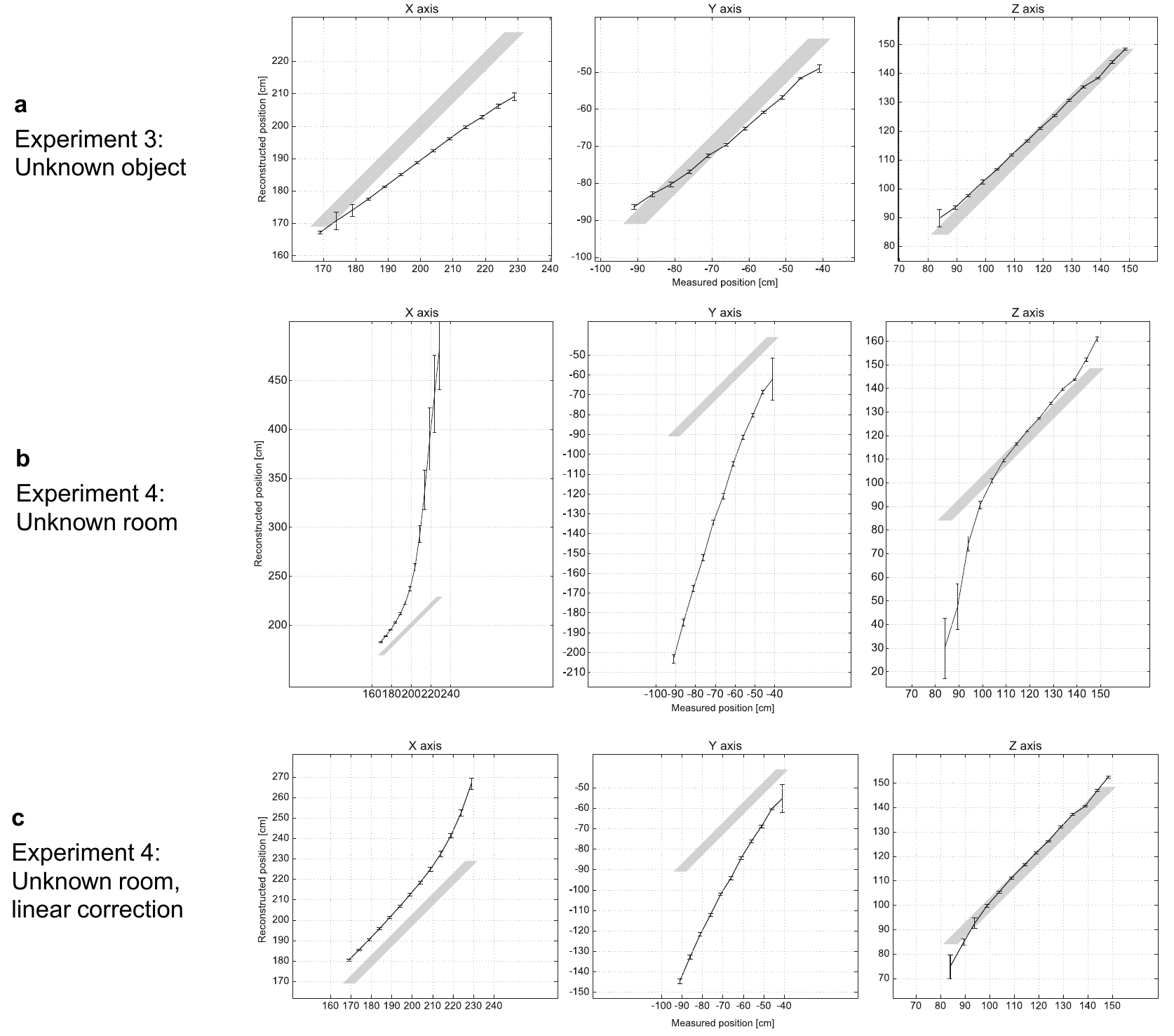}
\caption{
\label{fig:UnknownObjectTracking}
Tracking of an unknown object, or in an unknown room. {\bf\sffamily a,} Result of Experiment 3: Positional tracking as in Experiment 1, but with no knowledge about the object shape. We used a single oriented surface element for the light transport simulation. {\bf\sffamily b,} Result of Experiment 4: Positional tracking as in Experiment 1, but without subtracting the pre-calibrated room response. The estimated absolute position greatly deviated from the ground-truth position (shaded areas). {\bf\sffamily c,} Subtraction of a linear fit significantly reduced the tracking error and makes the tracking task feasible even in the absence of a background measurement. In all cases, the standard deviation (error bars) remained small, indicating that changes in position could still be robustly detected.
}
\end{figure}

To evaluate the method, we performed a series of experiments that are analysed in Fig.~\ref{fig:KnownObjectTracking} and \ref{fig:UnknownObjectTracking}. The physical object used in all experiments was a car silhouette cut from plywood and coated with white wall paint, shown in Fig.~\ref{fig:ObjectShape}(a). While our setup is able to handle arbitrary three-dimensional objects (as long as the convexity assumption is reasonable), this shape was two-dimensional for manufacturing and handling reasons.

For a given input image $\measuredImage$ and object shape, the cost function $f(\parameter)$ in Equation~(1) depends on three to six degrees of freedom that are being tracked. Fig.~\ref{fig:ObjectShape}(b) shows a slice of the function for translation in the XY-plane, with all other parameters fixed. Although the global minimum is located in an elongated, curved trough, only four to five iterations of the Levenberg-Marquardt algorithm are required for convergence from a random location in the tracking volume. In real-time applications, since position and rotation can be expected to change slowly over time, the optimization effort can be reduced to two to three iterations per frame by using the latest tracking result to initialize the solution for the next frame.

In {\bf Experiment 1}, we kept the object's orientation constant. We manually placed the object at various known locations in an 60\,cm$\times$50\,cm$\times$60\,cm working volume, and recorded 100 camera frames at each location. These frames differ in the amount of ambient light (mains flicker) and in the photon noise. For each frame, we initialized the estimated position to a random starting point in a cube of dimensions (30\,cm)$^3$ centered in the tracking volume, and refined the position estimate by minimizing the cost function (Equation~1). The results are shown in Fig.~\ref{fig:KnownObjectTracking}(a). From this experiment, we found positional tracking to be repeatable and robust to noise, with a sub-cm standard deviation for each position estimate. The root-mean-square distance to ground truth was measured at 4.8\,cm, 2.9\,cm and 2.4\,cm for movement along the X, Y and Z axis, respectively. This small systematic bias was likely caused by a known shortcoming of the image formation model, which does not account for occlusion of ambient light by the object.

In {\bf Experiment 2}, we kept the object at a (roughly) fixed location and rotated it by a range of $\pm$30$^\circ$ around the three coordinate axes using a pan-tilt-roll tripod with goniometers on all joints. Again, we recorded 100 frames per setting. We followed the same procedure as in the first experiment, except that this time we jointly optimized for all six degrees of freedom (position and orientation). The results are shown in Fig.~\ref{fig:KnownObjectTracking}(b). As expected, the rotation angles were tracked with higher uncertainty than the translational parameters. We identify two main sources for this uncertainty: the increased number of degrees of freedom and the pairwise ambiguity between X translation and Z rotation, and between Z translation and X rotation (Fig.~\ref{fig:DifferenceImages}). We recall that in the synthetic experiment, the effect of Y rotation was vanishingly small; here, the system tracked rotation around the Y axis about as robustly as the other axes. This unexpectedly positive result was probably owed to the strongly asymmetric shape of the car object. 

\begin{figure}
\centering
\includegraphics[width=\linewidth]{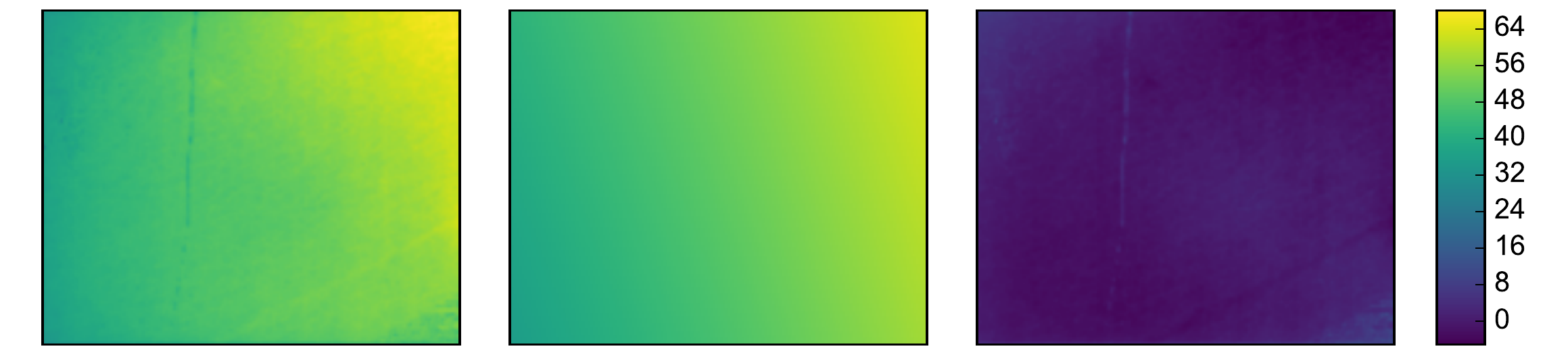}
\caption{
\label{fig:BackgroundLinearity}
Approximating the background term by a linear model. From left to right: background term $\widehat\backgroundImage$  obtained through calibration, linear approximation of $\widehat\backgroundImage$, residual background term after subtraction of linear component.
}
\end{figure}

So far, we assumed that the object's shape was known. Since this requirement cannot always be met, we dropped it in {\bf Experiment 3}. We used the data already captured using the car object for the first experiment, but performed the light transport simulation using a single oriented surface element instead of the detailed object model. Except for this simplification, we followed the exact same procedure as in Experiment 1 to track the now unknown object's position. The results are shown in Fig.~\ref{fig:UnknownObjectTracking}(a).
Despite a significant systematic shift introduced by the use of the simplified object model, the position recovery remained robust to noise and relative movement was still detected reliably.

The need for a measured background term $\widehat\backgroundImage$ can hinder the practical applicability of our approach as pursued so far. In {\bf Experiment 4}, we lifted this requirement. When omitting the term without any compensation, the tracking performance degraded significantly (Fig.~\ref{fig:UnknownObjectTracking}(b)). However, we observed that the background image, caused by distant scattering, was typically smooth and well approximated by a linear function $g(u,v)=au + bv + c$ in the image coordinates $u$ and $v$ (Fig.~\ref{fig:BackgroundLinearity}). We extended the tracking algorithm to fit such linear models to both input images $\measuredImage$ and $\simulatedImage{\parameter}$, and subtract the linear portions prior to evaluating the cost function (Eq.~2). This simple pre-processing step greatly reduced the bias in the tracking outcome and enabled robust tracking of object motion (Fig.~\ref{fig:UnknownObjectTracking}(c)) even in unknown rooms.

The supplementary material to this paper contains two videos, each showing a real-time tracking session (Session 1: translation only; Session 2: translation and rotation) using the described setup. A live view of the hidden scene is shown alongside screen output from the tracking software.  The average reconstruction rate during these tracking sessions was 10.2 frames per second (limited by the maximum capture rate of our camera-laser setup) for Session 1, and 3.7 frames per second (limited by computation) for Session 2. The 2-dimensional car model was represented by 502 surfels; the total compute time required for a single tracking step was 72.9\,ms for translation only, and 226.1\,ms for translation and rotation.

\paragraph{Discussion. }

In this work, we showed that the popular challenge of tracking an object around a corner does not inherently necessitate the use of time-of-flight technology. Rather, by formulating an optimization problem based on a simplistic image formation model, we demonstrated parametric object tracking in a room-sized scene with sub-cm repeatability, only using 2D images with a laser pointer as the light source. As our technique, in its current form, does not rely on temporally resolved measurements of any kind, it has the unique property of being scalable to very small scenes (down to the diffraction limit) as well as large scenes (sufficient laser power provided). We identify two main limiting factors to the performance of our technique: a systematic bias caused by shortcomings in the scene and light transport models, and high sensitivity to image noise when tracking object rotation. The adoption of advanced light transport models and noise reduction techniques will further improve the tracking quality. 
We note that the analysis-by-synthesis approach \emph{per se} is not limited to intensity imaging, but may form a valuable complement to other sensing modalities as well. For instance, a simple extension to the light transport model would enable it to accommodate time-of-flight imaging.
Like in all prior work, we assumed knowledge about the geometry and reflectance of a wall that receives light scattered by the unknown object. Thanks to recent progress in mobile mapping\cite{Puente20132127}, such data is already widely available for many application scenarios. We imagine a potential application of our technique to be in urban traffic safety, where the motion of vehicles and pedestrians is constrained to the ground plane and hence described by a small number of degrees of freedom.

\newpage


\section*{Methods}
\paragraph{Light transport simulation.} By assuming that all light has undergone exactly three reflections, we can efficiently simulate indirect illumination, with an overall computational complexity that is linear in the number of pixels and the number of surfels $n$. The geometry of this simulation is provided in Fig.1b. Each camera pixel observes a radiance value, $L$, leaving from a point on the wall, $p_W$, that, in turn, receives light reflected by the object's surfels. The portion contributed by the surfel of index $i\in \{1\dots n\}$ is the product of three reflectance terms, one per reflection event; and the geometric view factors known from radiative transfer\cite{cengel2014heat,Goral:1984:MIL:964965.808601}:
\begin{eqnarray}L_i:=&\rho_0\cdot f_s(p_L-p_S,p_i-p_S)&\textrm{(laser spot)}\\ \nonumber
&\cdot\,\,\frac{\left(n_S\circ\left(p_i-p_S\right)\right)\cdot \left(n_i\circ\left(p_S-p_i\right)\right)}{||p_S-p_i||_2^2} \cdot f_i(p_S-p_i,p_W-p_i)\cdot A_i&\textrm{(}i^\textrm{{th}}\textrm{ surfel)}\\ \nonumber
&\cdot\,\,\frac{\left(n_i\circ\left(p_W-p_i\right)\right)\cdot \left(n_W\circ\left(p_i-p_W\right)\right)}{||p_i-p_W||_2^2} \cdot f_W(p_i-p_W,p_C-p_W)&\textrm{(wall)},
\end{eqnarray}
where the operator \[
\normalizedDotProduct{\genericVectorA}{\genericVectorB}:=\begin{cases}
\frac{\transpose{\genericVectorA}\cdot\genericVectorB}{\norm{\genericVectorA}\cdot\norm{\genericVectorB}} & \mbox{if }\transpose{\genericVectorA}\cdot\genericVectorB>0\\
0 & \mbox{otherwise}
\end{cases}
\] 
denotes a normalized and clamped dot product as used in Lambert's cosine law.
Each line in Eq.~(2) models one of the three surface interactions. $n_S$, $n_i$ and $n_W$ are the normal vectors of laser spot, surfel and observed point on the wall, and $f_{\{S,i,W\}}(\omega_\mathrm{in},\omega_\mathrm{out})$ are the values of the corresponding bidirectional reflectance distribution functions (BRDF). The incident and outgoing direction vectors $\omega_\mathrm{in}$ and $\omega_\mathrm{out}$ that form the arguments to the BRDF are given by the scene geometry. In particular, the vectors $p_L$, $p_S$, $p_i$, $p_W$ and $p_C$ represent the positions of, in this order: the laser source, the laser spot on the wall, the $i^{\tiny\textrm{th}}$ surfel, the observed point on the wall, and the camera (center of projection). $A_i$ is the area of the $i^{\tiny\textrm{th}}$ surfel, and $\rho_0$ a constant factor that subsumes laser power and the light efficiencies of lens and sensor. This factor is cancelled out by the projection performed in the cost function Eq.~(1), so we set it to $\rho_0=1$ in simulation. The total pixel value is simply computed by summing Eq.~(2) over all surfels:
\begin{equation}
L_{\mathrm{total}} := \sum_{i=1}^n L_i
\label{eq:totalirradiance}
\end{equation}
This summation neglects mutual shadowing or inter-reflection between surfels, an approximation that is justifiable for flat or mostly convex objects. 
For lack of measured material BRDFs, we further assume all surfaces to be of diffuse (Lambertian) reflectance such that $f_{\{S,i,W\}}:=\mathrm{const}=1$, again making use of the fact that the cost function  Eq.~(1) is invariant under such global scaling factors. If available, more accurate BRDF models as well as object and wall textures can be included at a negligible computational cost.

\paragraph{Capture devices.}
Our image source was a Xenics Xeva-1.7-320 camera, sensitive in the near-infrared range (900\,nm--1,700\,nm), with a resolution of 320$\times$256 pixels at 14 bits per pixel. We used an exposure time of 20\,ms.
The laser source (1\,W at 1.550\,nm) was a fiber-coupled
laser diode of type SemiNex 4PN-108 driven by an Analog Technologies ATLS4A201D laser diode driver and equipped with a USB interface trigger input. On the output side of the fiber, we fed the collimated beam through a narrow tube with absorbing walls to reduce stray light.

A desktop PC with an NVIDIA GeForce GTX 780 GPU, 32GB of RAM and an Intel Core
i7-4930K CPU controlled the devices and performed the reconstruction.

\paragraph{Measurement routine and image pre-processing.}

After calibrating the camera's gain factors and fixed pattern noise using vendor tools, we assumed that all pixels had the same linear response. All images were downsampled to half the resolution (160$\times$128 pixels) prior to further processing. Due to the diffuse reflections, the measurements do not contain any high-frequent information apart from noise, thus moderate down sampling is a safe way to improve the performance of the later reconstruction.

The images measured by the camera are composed of several contributions, each represented by a vector of pixel-wise contributions:  \emph{ambient} light not originating from the laser, $\ambientImage$; laser light scattered by static \emph{background} objects present in the scene, $\backgroundImage$; and laser light scattered by the dynamic \emph{object}, $\objectImage
$. All measured images are further affected by noise, the main sources being photon counting noise and signal-independent read noise. We assume the scene to remain stationary at least during short time intervals 
between successive captures. Further assuming the spatial extent of the object to be small, shadowing of $\ambientImage$ and $\backgroundImage$ by the object, as well as ambient light reflected by the object, can be neglected. By turning the laser on and off, and inserting and removing the object, the described kind of setup can capture the following  combinations of these light contributions:

\begin{tabbing}
Laser off (0), object absent (0):\qquad\=$\cameraImage_{00}
=\ambientImage + \mathrm{noise}$\\
Laser on (1), object absent (0):\>$\cameraImage_{10}
=\ambientImage+\backgroundImage + \mathrm{noise}$\\
Laser off (0), object present (1):\>$\cameraImage_{01}
=\ambientImage + \mathrm{noise}$\\
Laser on (1), object present (1):\>$\cameraImage_{11}
=\ambientImage+\backgroundImage+\objectImage
 + \mathrm{noise}$
\end{tabbing}

The input image to the reconstruction algorithm, $\measuredImage$, was obtained as the difference of images captured in quick succession with and without laser illumination. Additionally, we subtracted a calibration measurement containing light reflected by the background:
\begin{eqnarray}
\measuredImage
:=\cameraImage_{11}
-\cameraImage_{01}
-\widehat\backgroundImage\approx\objectImage
 + \mathrm{noise},
\end{eqnarray}
The addition or subtraction of two input images increases the noise magnitude by a factor of about $\sqrt{2}$. The background estimate $\widehat\backgroundImage$ was captured with the object removed by recording difference images with and without laser illumination. We averaged $n=300$ such difference images to reduce noise in the background estimate:
\begin{eqnarray}
\widehat\backgroundImage:=\frac 1n\sum_{i=0}^n\left(\cameraImage_{10}^{(i)}-\cameraImage_{00}^{(i)}\right)\stackrel{\,\,n>\!\!>1\,\,}{\approx}\backgroundImage
\end{eqnarray}
%



\section*{References}
\bibliographystyle{naturemag}

\begin{thebibliography}{10}
\expandafter\ifx\csname url\endcsname\relax
  \def\url#1{\texttt{#1}}\fi
\expandafter\ifx\csname urlprefix\endcsname\relax\def\urlprefix{URL }\fi
\providecommand{\bibinfo}[2]{#2}
\providecommand{\eprint}[2][]{\url{#2}}

\bibitem{Abramson:78}
\bibinfo{author}{Abramson, N.}
\newblock \bibinfo{title}{Light-in-flight recording by holography}.
\newblock \emph{\bibinfo{journal}{Optics Letters}}
  \textbf{\bibinfo{volume}{3}}, \bibinfo{pages}{121--123}
  (\bibinfo{year}{1978}).
\newblock \urlprefix\url{http://ol.osa.org/abstract.cfm?URI=ol-3-4-121}.

\bibitem{Velten:2011}
\bibinfo{author}{Velten, A.}, \bibinfo{author}{Raskar, R.} \&
  \bibinfo{author}{Bawendi, M.}
\newblock \bibinfo{title}{Picosecond camera for time-of-flight imaging}.
\newblock In \emph{\bibinfo{booktitle}{Imaging and Applied Optics}},
  \bibinfo{pages}{IMB4} (\bibinfo{publisher}{Optical Society of America},
  \bibinfo{year}{2011}).
\newblock
  \urlprefix\url{http://www.osapublishing.org/abstract.cfm?URI=ISA-2011-IMB4}.

\bibitem{Velten:2012:Recovering}
\bibinfo{author}{Velten, A.} \emph{et~al.}
\newblock \bibinfo{title}{Recovering three-dimensional shape around a corner
  using ultrafast time-of-flight imaging}.
\newblock \emph{\bibinfo{journal}{Nature Communications}}
  \textbf{\bibinfo{volume}{3}}, \bibinfo{pages}{745} (\bibinfo{year}{2012}).

\bibitem{Heide:2014}
\bibinfo{author}{Heide, F.}, \bibinfo{author}{Xiao, L.},
  \bibinfo{author}{Heidrich, W.} \& \bibinfo{author}{Hullin, M.~B.}
\newblock \bibinfo{title}{Diffuse mirrors: {3D} reconstruction from diffuse
  indirect illumination using inexpensive time-of-flight sensors}.
\newblock \emph{\bibinfo{journal}{IEEE Conf. on Computer Vision and Pattern
  Recognition (CVPR)}}  (\bibinfo{year}{2014}).

\bibitem{Laurenzis:2014}
\bibinfo{author}{Laurenzis, M.} \& \bibinfo{author}{Velten, A.}
\newblock \bibinfo{title}{Nonline-of-sight laser gated viewing of scattered
  photons}.
\newblock \emph{\bibinfo{journal}{Optical Engineering}}
  \textbf{\bibinfo{volume}{53}}, \bibinfo{pages}{023102--023102}
  (\bibinfo{year}{2014}).

\bibitem{Gariepy:2016}
\bibinfo{author}{Gariepy, G.}, \bibinfo{author}{Tonolini, F.},
  \bibinfo{author}{Henderson, R.}, \bibinfo{author}{Leach, J.} \&
  \bibinfo{author}{Faccio, D.}
\newblock \bibinfo{title}{Detection and tracking of moving objects hidden from
  view}.
\newblock \emph{\bibinfo{journal}{Nature Photonics}}
  \textbf{\bibinfo{volume}{10}} (\bibinfo{year}{2016}).

\bibitem{Sen:2005:DP}
\bibinfo{author}{Sen, P.} \emph{et~al.}
\newblock \bibinfo{title}{{Dual Photography}}.
\newblock \emph{\bibinfo{journal}{ACM Transactions on Graphics (Proc.
  SIGGRAPH)}} \textbf{\bibinfo{volume}{24}}, \bibinfo{pages}{745--755}
  (\bibinfo{year}{2005}).

\bibitem{Steinvall:2011}
\bibinfo{author}{Steinvall, O.}, \bibinfo{author}{Elmqvist, M.} \&
  \bibinfo{author}{Larsson, H.}
\newblock \bibinfo{title}{See around the corner using active imaging}.
\newblock \emph{\bibinfo{journal}{Proc. SPIE}} \textbf{\bibinfo{volume}{8186}},
  \bibinfo{pages}{818605--818605--17} (\bibinfo{year}{2011}).
\newblock \urlprefix\url{http://dx.doi.org/10.1117/12.893605}.

\bibitem{katz2012looking}
\bibinfo{author}{Katz, O.}, \bibinfo{author}{Small, E.} \&
  \bibinfo{author}{Silberberg, Y.}
\newblock \bibinfo{title}{Looking around corners and through thin turbid layers
  in real time with scattered incoherent light}.
\newblock \emph{\bibinfo{journal}{Nature Photonics}}
  \textbf{\bibinfo{volume}{6}}, \bibinfo{pages}{549--553}
  (\bibinfo{year}{2012}).

\bibitem{Katz:2014}
\bibinfo{author}{Katz, O.}, \bibinfo{author}{Heidmann, P.},
  \bibinfo{author}{Fink, M.} \& \bibinfo{author}{Gigan, S.}
\newblock \bibinfo{title}{Non-invasive single-shot imaging through scattering
  layers and around corners via speckle correlations}.
\newblock \emph{\bibinfo{journal}{Nature Photonics}}
  \textbf{\bibinfo{volume}{8}}, \bibinfo{pages}{784--790}
  (\bibinfo{year}{2014}).

\bibitem{Sume:2011}
\bibinfo{author}{Sume, A.} \emph{et~al.}
\newblock \bibinfo{title}{Radar detection of moving targets behind corners}.
\newblock \emph{\bibinfo{journal}{Geoscience and Remote Sensing, IEEE
  Transactions on}} \textbf{\bibinfo{volume}{49}}, \bibinfo{pages}{2259--2267}
  (\bibinfo{year}{2011}).

\bibitem{Adib:2013}
\bibinfo{author}{Adib, F.} \& \bibinfo{author}{Katabi, D.}
\newblock \bibinfo{title}{See through walls with wifi!}
\newblock \emph{\bibinfo{journal}{SIGCOMM Comput. Commun. Rev.}}
  \textbf{\bibinfo{volume}{43}}, \bibinfo{pages}{75--86}
  (\bibinfo{year}{2013}).
\newblock \urlprefix\url{http://doi.acm.org/10.1145/2534169.2486039}.

\bibitem{Adib:2015}
\bibinfo{author}{Adib, F.}, \bibinfo{author}{Hsu, C.-Y.}, \bibinfo{author}{Mao,
  H.}, \bibinfo{author}{Katabi, D.} \& \bibinfo{author}{Durand, F.}
\newblock \bibinfo{title}{Capturing the human figure through a wall}.
\newblock \emph{\bibinfo{journal}{ACM Trans. Graph. (Proc. SIGGRAPH Asia)}}
  \textbf{\bibinfo{volume}{34}}, \bibinfo{pages}{219:1--219:13}
  (\bibinfo{year}{2015}).
\newblock \urlprefix\url{http://doi.acm.org/10.1145/2816795.2818072}.

\bibitem{Buttafava:2015}
\bibinfo{author}{Buttafava, M.}, \bibinfo{author}{Zeman, J.},
  \bibinfo{author}{Tosi, A.}, \bibinfo{author}{Eliceiri, K.} \&
  \bibinfo{author}{Velten, A.}
\newblock \bibinfo{title}{Non-line-of-sight imaging using a time-gated single
  photon avalanche diode}.
\newblock \emph{\bibinfo{journal}{Optics Express}}
  \textbf{\bibinfo{volume}{23}}, \bibinfo{pages}{20997--21011}
  (\bibinfo{year}{2015}).

\bibitem{Pan:2009}
\bibinfo{author}{Pan, X.}, \bibinfo{author}{Sidky, E.~Y.} \&
  \bibinfo{author}{Vannier, M.}
\newblock \bibinfo{title}{Why do commercial {CT} scanners still employ
  traditional, filtered back-projection for image reconstruction?}
\newblock \emph{\bibinfo{journal}{Inverse Problems}}
  \textbf{\bibinfo{volume}{25}}, \bibinfo{pages}{123009}
  (\bibinfo{year}{2009}).
\newblock \urlprefix\url{http://stacks.iop.org/0266-5611/25/i=12/a=123009}.

\bibitem{Kadambi:2016:OIT:2882845.2836164}
\bibinfo{author}{Kadambi, A.}, \bibinfo{author}{Zhao, H.},
  \bibinfo{author}{Shi, B.} \& \bibinfo{author}{Raskar, R.}
\newblock \bibinfo{title}{Occluded imaging with time-of-flight sensors}.
\newblock \emph{\bibinfo{journal}{ACM Transactions on Graphics}}
  \textbf{\bibinfo{volume}{35}}, \bibinfo{pages}{15:1--15:12}
  (\bibinfo{year}{2016}).
\newblock \urlprefix\url{http://doi.acm.org/10.1145/2836164}.

\bibitem{Abramson:83:LIF}
\bibinfo{author}{Abramson, N.}
\newblock \bibinfo{title}{Light-in-flight recording: high-speed holographic
  motion pictures of ultrafast phenomena}.
\newblock \emph{\bibinfo{journal}{Applied optics}}
  \textbf{\bibinfo{volume}{22}}, \bibinfo{pages}{215--232}
  (\bibinfo{year}{1983}).

\bibitem{Quercioli:85}
\bibinfo{author}{Quercioli, F.} \& \bibinfo{author}{Molesini, G.}
\newblock \bibinfo{title}{White light-in-flight holography}.
\newblock \emph{\bibinfo{journal}{Applied Optics}}
  \textbf{\bibinfo{volume}{24}}, \bibinfo{pages}{3406--3415}
  (\bibinfo{year}{1985}).
\newblock \urlprefix\url{http://ao.osa.org/abstract.cfm?URI=ao-24-20-3406}.

\bibitem{goda2009}
\bibinfo{author}{Goda, K.}, \bibinfo{author}{Tsia, K.~K.} \&
  \bibinfo{author}{Jalali, B.}
\newblock \bibinfo{title}{Serial time-encoded amplified imaging for real-time
  observation of fast dynamic phenomena}.
\newblock \emph{\bibinfo{journal}{Nature}} \textbf{\bibinfo{volume}{458}},
  \bibinfo{pages}{1145--1149} (\bibinfo{year}{2009}).

\bibitem{gariepy2015single}
\bibinfo{author}{Gariepy, G.} \emph{et~al.}
\newblock \bibinfo{title}{Single-photon sensitive light-in-flight imaging}.
\newblock \emph{\bibinfo{journal}{Nature Communications}}
  \textbf{\bibinfo{volume}{6}} (\bibinfo{year}{2015}).

\bibitem{Heide:2013:LBT}
\bibinfo{author}{Heide, F.}, \bibinfo{author}{Hullin, M.~B.},
  \bibinfo{author}{Gregson, J.} \& \bibinfo{author}{Heidrich, W.}
\newblock \bibinfo{title}{Low-budget transient imaging using photonic mixer
  devices}.
\newblock \emph{\bibinfo{journal}{ACM Transactions on Graphics (Proc.
  SIGGRAPH)}} \textbf{\bibinfo{volume}{32}}, \bibinfo{pages}{45:1--45:10}
  (\bibinfo{year}{2013}).

\bibitem{kadambi2013coded}
\bibinfo{author}{Kadambi, A.} \emph{et~al.}
\newblock \bibinfo{title}{Coded time of flight cameras: sparse deconvolution to
  address multipath interference and recover time profiles}.
\newblock \emph{\bibinfo{journal}{ACM Trans. Graph. (Proc. SIGGRAPH Asia)}}
  \textbf{\bibinfo{volume}{32}}, \bibinfo{pages}{167} (\bibinfo{year}{2013}).

\bibitem{peters-2015-fast-transient}
\bibinfo{author}{Peters, C.}, \bibinfo{author}{Klein, J.},
  \bibinfo{author}{Hullin, M.~B.} \& \bibinfo{author}{Klein, R.}
\newblock \bibinfo{title}{Solving trigonometric moment problems for fast
  transient imaging}.
\newblock \emph{\bibinfo{journal}{ACM Trans. Graph. (Proc. SIGGRAPH Asia)}}
  \textbf{\bibinfo{volume}{34}}, \bibinfo{pages}{220:1--220:11}
  (\bibinfo{year}{2015}).

\bibitem{marquardt1963algorithm}
\bibinfo{author}{Marquardt, D.~W.}
\newblock \bibinfo{title}{An algorithm for least-squares estimation of
  nonlinear parameters}.
\newblock \emph{\bibinfo{journal}{Journal of the society for Industrial and
  Applied Mathematics}} \textbf{\bibinfo{volume}{11}},
  \bibinfo{pages}{431--441} (\bibinfo{year}{1963}).

\bibitem{ceres-solver}
\bibinfo{author}{Agarwal, S.}, \bibinfo{author}{Mierle, K.} \&
  \bibinfo{author}{{Others~[sic]}}.
\newblock \bibinfo{title}{Ceres solver}.
\newblock \bibinfo{howpublished}{\url{http://ceres-solver.org}}
  (\bibinfo{year}{2015}).

\bibitem{Puente20132127}
\bibinfo{author}{Puente, I.}, \bibinfo{author}{Gonz\'alez-Jorge, H.},
  \bibinfo{author}{Mart\'inez-S\'anchez, J.} \& \bibinfo{author}{Arias, P.}
\newblock \bibinfo{title}{Review of mobile mapping and surveying technologies}.
\newblock \emph{\bibinfo{journal}{Measurement}} \textbf{\bibinfo{volume}{46}},
  \bibinfo{pages}{2127 -- 2145} (\bibinfo{year}{2013}).

\bibitem{cengel2014heat}
\bibinfo{author}{{\c{C}}engel, Y.} \& \bibinfo{author}{Ghajar, A.}
\newblock \emph{\bibinfo{title}{Heat and Mass Transfer: Fundamentals and
  Applications}} (\bibinfo{publisher}{McGraw-Hill Education},
  \bibinfo{year}{2014}).

\bibitem{Goral:1984:MIL:964965.808601}
\bibinfo{author}{Goral, C.~M.}, \bibinfo{author}{Torrance, K.~E.},
  \bibinfo{author}{Greenberg, D.~P.} \& \bibinfo{author}{Battaile, B.}
\newblock \bibinfo{title}{Modeling the interaction of light between diffuse
  surfaces}.
\newblock \emph{\bibinfo{journal}{SIGGRAPH Computer Graphics}}
  \textbf{\bibinfo{volume}{18}}, \bibinfo{pages}{213--222}
  (\bibinfo{year}{1984}).
\newblock \urlprefix\url{http://doi.acm.org/10.1145/964965.808601}.

\end{thebibliography}


\final{
\begin{addendum}
 \item This work was supported by the X-Rite Chair for Digital Material Appearance, and the German Research Foundation (HU 2273/2-1). We thank Frank Christnacher and Emmanuel Bacher for their help with the setup and discussion.
\item[Author Contributions] The method was conceived by M.B.H. and J.K. All authors took part in designing the experiments. J.K. implemented and optimized the system based on prototypes by M.B.H. and J.M. M.B.H. and J.K. wrote the paper.
\item[Competing Interests] The authors declare that they have no
competing financial interests.
\item[Correspondence] Correspondence and requests for materials
should be addressed to M.B.H.~(email: hullin@cs.uni-bonn.de).
\end{addendum}
\newpage
}

\end{document}